# TF-IDF vs Word Embeddings for Morbidity Identification in Clinical Notes: An Initial Study

Danilo Dess`ı[1] [0000−0003−3843−3285], Rim Helaoui[2] [0000−0001−6915−8920], Vivek Kumar[1] [0000−0003−3958−4704], Diego Reforgiato Recupero[1] [0000−0001−8646−6183], and Daniele Riboni[1] [0000−0002−0695−2040]

[1] University of Cagliari, Cagliari, Italy
`{danilo_dessi, vivek.kumar, diego.reforgiato, riboni}@unica.it`
[2] Philips Research, Eindhoven, Netherlands
`rim.helaoui@philips.com`

**Abstract.** Today, we are seeing an ever-increasing number of clinical notes that contain clinical results, images, and textual descriptions of pa-tient's health state. All these data can be analyzed and employed to cater novel services that can help people and domain experts with their com-mon healthcare tasks. However, many technologies such as Deep Learn-ing and tools like Word Embeddings have started to be investigated only recently, and many challenges remain open when it comes to healthcare domain applications. To address these challenges, we propose the use of Deep Learning and Word Embeddings for identifying sixteen morbidity types within textual descriptions of clinical records. For this purpose, we have used a Deep Learning model based on Bidirectional Long-Short Term Memory (LSTM) layers which can exploit state-of-the-art vector representations of data such as Word Embeddings. We have employed pre-trained Word Embeddings namely GloVe and Word2Vec, and our own Word Embeddings trained on the target domain. Furthermore, we have compared the performances of the deep learning approaches against the traditional tf-idf using Support Vector Machine and Multilayer per-ceptron (our baselines). From the obtained results it seems that the lat-ter outperform the combination of Deep Learning approaches using any word embeddings. Our preliminary results indicate that there are specific features that make the dataset biased in favour of traditional machine learning approaches.

## 1 Introduction

In these years we are seeing an increment of life expectancy that has also in-creased the risk of long-term diseases such as cancer, diabetes, mental health condition, and other chronic health threats [22, 10, 3, 21]. Also, one more disad-vantage with long life expectancy is that people can be affected by more than one disease at a time, increasing the risk of substandard health quality. For instance, a person suffering from long term diabetes have higher chances of hypertension, high cholesterol levels, arteries or veins blockage. The World Health Organiza-tion report [17] states that, in a developed country, over 40% of the population is exposed to at least one long-term health condition including all ages, and 25% percent of the population suffers with multi-morbidity. Furthermore, the report emphasizes that the high rate of multi-morbidity is directly proportional to the middle and low-income countries as they do not have funds that should be invested for enhancing primary care of population [2]. All this medical information is being monitored, and with the advent of Information Technology (IT) services, a lot of clinical data are continuously being stored within clinical reports, which might be employed to provide novel healthcare services world-wide, overcoming issues related to the social or economic condition of people. Clinical reports may contain various information in the form of numbers (e.g., laboratory results), images (e.g., x-ray), or medical descriptions (e.g., surgical descriptions) that may be used to create content-based services. However, the whole amount of data is challenging to be analyzed and employed by humans to provide healthcare services, and the development of computer-based systems to deal with it has recently gained the attention of the scientific community. For example, textual data of clinical reports have been explored in tasks such as classification [4], clustering [12], and recommendation [8]. Although state-of-the-art research in this direction has already provided important outcomes, many challenges still remain open [5]. Methods based on Deep Learning models and advanced Word Embedding representations have recently proven to be state-of-the-art for many tasks, but their use within many healthcare problems have not been investigated yet. Therefore, in this paper we investigate the use of a Deep Learning model with various types of Word Embeddings in order to address the problem of morbidity detection related to the obesity disease in textual clinical notes. We performed our research study on the n2c2[3] dataset released for the i2b2[4] obesity and co-morbidity detection challenge in 2008, and compared our results against two baseline approaches where the Term Frequency — In-verse Document Frequency (TF-IDF) was used to represent clinical notes. More precisely, the contributions of our paper are:

- We provide a Deep Learning model using Word Embeddings for performing multi-morbidity detection within clinical notes.
- We analyze three types of Word Embeddings (for the Deep Learning approaches) and the TF-IDF (for the baselines) representation for modelling the knowledge of clinical notes.
- We compare the proposed deep learning approaches against two machine learning methods by using k-fold cross-validation.
- We found out a very high f-measure score of the machine learning baselines that beats those of the deep learning approaches indicating the occurrences of representative tokens highly connected with the presence of each morbidity class.
– We make available all the sources code used for our experiments by a GitHub repository[5].

---

[3] https://n2c2.dbmi.hms.harvard.edu/

[4] https://www.i2b2.org/NLP/Obesity/

The remaining of this manuscript is organized as it follows: Section 2 discusses the role of Deep Learning and Natural Language Processing (NLP) techniques within the healthcare domain. Section 3 describes the dataset we have used, its contents and how it was created. Section 4 introduces the methodology we have applied, and the feature engineering approaches we have designed for the Deep Learning task. Section 5 presents and discusses the results. Finally, Section 6 draws some conclusions of our preliminary investigation and describes problems to solve and future research challenges where we are headed.

## 2 Related Work

These days Artificial Intelligence (AI) and its sub fields such as Deep Learning, Text Mining, and more in general Machine Learning, are playing a significant role in clinical decision making and understanding, automatic disease diagnosis, and therapy assistance [15]. Deep Learning applications in healthcare are con-tributing with relevant improvements in many fields such as analyzing the blood samples, detecting heart problems, detecting tumours, and so on [19]. Moreover, the high quality performances of Deep Learning models for healthcare issues has raised positive discussions and interests within the AI community. However, the use of Deep Learning technologies for the purpose of detecting multi-morbidity in clinical notes has not been deeply analyzed yet. For example, a relevant re-cent work [13] uses Negative Matrix Factorisation (NMF) for simultaneously mining disease clusters in order to detect relations that exist between morbidity patterns. Authors demonstrated how the temporal characteristics of the disease clusters can help mining multi-morbidity networks and generating new hypothe-ses for the emergence of various morbidity patterns over time. However, no Deep Learning methods have been investigated to try to uncover morbidity patterns. Other works only rely on NLP techniques [23]. For example, authors in [11] present a method called FREGEX which is based on regular expressions to ex-tract features from written clinical notes. The use of Deep Learning models for discovering multi-morbidity linked to the obesity disease has been recently inves-tigated by [24]. The Deep Learning model used in this work presents two layers, a Convolutional Neural Network (CNN) layer and a Max Pooling layer, that were fed by using both word and entity embeddings, and allowed the authors to improve the results that were obtained during the i2b2 obesity challenge in 2008 especially for the intuitive classification task. However, they investigated only CNN as the main layer of their model and, therefore, the use of Long-Short Term Memory (LSTM) still needs to be explored. In addition, in literature there is a strong evidence that Bidirectional layers can infer relevant characteristic from clinical notes [14]. Hence, in this work we investigate the use of Bidirec-tional LSTM layers for the task of multi-morbidity detection. Moreover, we try to understand whether state-of-the-art Word Embeddings available in literature can better represent the knowledge of clinical notes for the same task.

---
[5] https://github.com/vsrana-ai/SmartPhil

## 3 Dataset Description

The dataset used for this work is n2c2 obesity data. The n2c2 dataset contains test and training documents of patients clinical records. The dataset was com-pletely anonymized by replacing personal and sensitive information of patients with surrogates. The morbidity classes in the dataset are *Asthma, CAD, CHF, Depression, Diabetes, Gallstones, GERD, Gout, Hypercholesterolemia, Hyperten-sion, Hypertriglyceridemia, OA, Obesity, OSA, PVD, and Venous Insufficiency*. Each clinical note within the dataset is associated with two types of labels: *intuitive and textual*. *Textual* labels indicate if there is a clear evidence that a clinical note presents a specific morbidity. On the other hand, *intuitive* means that there has been a domain expert (e.g., a physician) who, by reading the clinical note, was able to infer that the clinical state of a patient can suggest the presence of the target morbidity. The labels can assume a value in *{Y, N, U, Q}*, where "Y" means *yes, the patient has the morbidity*, "N" means *no, the patient does not have the morbidity*, "U" means *the morbidity is not mentioned in the record*, and "Q" stands for *questionable whether the patient has the morbidity*. When more domain experts disagreed on the morbidity for a specific clinical note, the label can occasionally assumes the values "Q" or "U". However, we only consider clin-ical notes that clearly showed a morbidity with labels that valued "Y" or "N". More specifically, given the set of all clinical notes, let us say $M$, and an input morbidity class $c$, we first selected all clinical notes which had "Y" or "N" as textual label for the morbidity class c thus building the set N. Then, we added to the set N all clinical notes in $M - N$ that had "Y" or "N" as *intuitive* label, yielding the set N'.

The set N' was used to create an unique dataset for each morbidity class c. In doing so, it is straightforward to apply binary classifiers for detecting each different morbidity in clinical notes. As final step for the dataset preparation, we have merged in a unique dataset all the notes coming from the training and test set of the original data. This was necessary in order to perform our experiments by using the k fold cross-validation approach. The size of each dataset, and the number of positive and negative labels within them is reported in Table 1.

## 4 Methodology

In this section we briefly describe which pre-processing steps have been per-formed on the input texts. Then, we describe the Deep Learning model and the data representations we employed to uncover the knowledge from our data. Finally, we report details about the experimental setup we designed.

Table 1. The size of dataset and the number of positive and negative labels for each binary morbidity class.

| Morbidity | Clinical notes | Positive | Negative |
|---|---|---|---|
| *Asthma* | 952 | 144 | 808 |
| *CAD* | 942 | 572 | 370 |
| *CHF* | 904 | 442 | 462 |
| *Depression* | 968 | 224 | 744 |
| *Diabetes* | 980 | 688 | 292 |
| *Gallstones* | 996 | 174 | 822 |
| *GERD* | 858 | 196 | 662 |
| *Gout* | 1,004 | 126 | 878 |
| *Hypercholesterolemia* | 876 | 496 | 380 |
| *Hypertension* | 942 | 176 | 766 |
| *Hypertriglyceridemia* | 976 | 54 | 922 |
| *OA* | 934 | 202 | 732 |
| *Obesity* | 930 | 414 | 516 |
| *OSA* | 994 | 140 | 854 |
| *PVD* | 938 | 140 | 798 |
| *Venous Insufficiency* | 858 | 62 | 796 |

## 4.1 Problem Statement

The problem we dealt with was a multi-class multi-label classification problem. We addressed the problem by using binary classifiers for the target classes of our dataset. In doing so we were able to investigate which classes were more chal-lenging for automatically inferring the knowledge about morbidity from textual resources. More precisely, given a target morbidity class c representing a morbid-ity and a clinical note text t, our purpose was to infer a function $\gamma(t, c) \rightarrow l$ where l is a binary label that can only assume values in {0, 1}. The label l = 0 means that the target morbidity c is not associated with the clinical note, whereas l = 1 means that from t that morbidity comes up.

## 4.2 Pre-processing

In order to prepare our textual data, we have performed the following steps:

1. We transformed our texts in lower case, so that the same word written in different cases (e.g., *Obesity* and *obesity*) could be represented by the same string (i.e., *obesity*).
2. We tokenized our texts and built a function *f*, where for each word *w* the function *f* was able to associate an integer index *i*.
3. We encoded our input texts using bag of word representations into integer numerical representations. For example, consider the sentence *s* "the patient has the diabetes" and a function f that maps "the" to "5", "patient" to "34", "has" to "10", "diabetes" to "87". Then, the integer-encoded sentence $s_{encoded}$ is [5, 34, 10, 5, 87].

4. Because each clinical note might be too long, and therefore difficult to be represented and potentially causing problems with the training step as there are not many notes, we limited the number of tokens for each text. It has been computed as the sum between the average and the standard deviation of the number of tokens each input text had. For example, imagine to have four texts with 25, 39, 44, and 80 tokens respectively. Then, the average length is *avg* = (25 + 39 + 44 + 80)/4 = 47.00 and the standard deviation *std* = 20.29, hence, the length that our method considers is 47 + 20 = 67. Another alternative that we are already tackling is to break a long clinical notes in more notes with the same annotations.

In addition, as far as the TF-IDF is concerned, we performed the following steps [16, 7] already adopted on this domain in literature:

1. We removed all stopwords by using the NLTK[6] library.
2. We remove punctuation and numerical values.
3. We have used the maximum number of features max features (padding wher-ever necessary) to design our TF-IDF matrix, so that no information is lost while creating the matrix. For our dataset, if the number of clinical reports is m and the overall number of distinct tokens is n, then the TF-IDF matrix is of dimension *{mxn}*, where *n = max f eatures.*

### 4.3 Data Modelling

Within our work, we used Word Embeddings and TF-IDF as representation models of input data. Word Embeddings are distributed representations that model words properties in vectors of real numbers and capture syntactic features and semantic word relationships. They have shown to be suitable to model the knowledge in many domains. On the other hand, TF-IDF was the most useful representation method for textual data and, therefore, it remains the baseline approach for any innovative method for text classification in various domains. Hence, the data model representations we used within our study are:

– Pre-trained Word2Vec [18]. The *Word2Vec* algorithm aims to detect the meaning and semantic relations by studying co-occurrences between words in a given corpus. Within our work we employed the Word Embeddings[7] of size 300 generated on google news.
– Domain-trained Word2Vec. These domain-trained Word Embeddings are trained with the same algorithm [18] on our morbidity dataset. Train-ing this kind of embeddings has shown advantages because they can embed semantics about jargons and specific terms of the target domain. Moreover, using Word Embeddings trained on the target domain avoids the problem of having out-of-vocabulary words since the less frequent words also have a vector representation. We generated these Word Embeddings of size 300 by using the gensim[8] library with 10 epochs and a window size of 5.

---

[6] https://www.nltk.org/
[7] https://code.google.com/archive/p/word2vec/
[8] https://radimrehurek.com/gensim/

– GloVe [20]. *GloVe* generator algorithm was proposed by the Stanford com-munity in 2014. This algorithm adopts a statistics-based matrix in order to represent how frequent words appear in a given context, and computes vectors scores based on co-occurrences of words within contexts. Within our work, we used the *GloVe6B* Word Embeddings of size 300.
– TF-IDF. It is a technique of data modelling that computes a weight for each word which indicates the importance of the word for a given document within a corpus. TF defines the occurrence of a word w in a document d. IDF measures the rarity of a word w in the whole document. Equation 1 shows the TF-IDF formula where $c^w_i$ is the number of occurrences of the word *w* in the *i-th* document $d_i$, $|d_i|$ is the size of the document expressed as number of words, N is the number of documents in the collection, and $n^w$ is the number of documents where the word w occurs at least once. TF-IDF values are usually normalized in the range [0,1].

$$TF-IDF(w, d_i) = \frac{c^w_i}{|d_i|} \cdot \log \frac{N}{n^w} \quad (1)$$

## 4.4 The Deep Learning and TF-IDF Models

Our Deep Learning model is depicted in Figure 1. The model is aimed at perform-ing binary classification and it takes inspiration from [6, 1], where Bidirectional Long Short Term Memory (BiLSTM) layers fed by Word Embedding representa-tions were able to well represent the knowledge of target domains and obtained state-of-the-art results. Differently from these previous models, our model can be changed to parse different data representations that we decided to employ. More specifically, in Figure 1 the model has an Embeddings layer as input layer. Embeddings layer takes the input Word Embeddings of size 300 and accepts integer-encoded vectors. More specifically, it organizes the input with a matrix with size (N × M) where N is the number of encoded texts, and M is the max-imum number of tokens considered for each text. Then, it loads an embeddings matrix that is used to link the indexes of encoded texts to vector representations. The output of the Embeddings layer is given to two hidden layers that implement BiLSTM neural networks. LSTM is a particular kind of recurrent neural network that is able to store the history of the input data and has already proven to be able to find patterns in data where the sequence of the information matters. By using the bidirectional version, the model is able to learn from the input data both backward and forward patterns that might be not detected parsing the data in just one direction. The results of backward LSTM and forward LSTM are combined into a unique result. Finally, the last layer of the model is a fully-connected Dense layer whose purpose is to predict the binary label. To compare our approach against a non deep learning model we have used Support Vector Machine and Multilayer Perceptron in combination with the TF-IDF, and the steps of the adopted process are shown in Figure 2. We have applied the steps 1 and 2 of pre-processing explained in Section 4.2, followed by removing the stop words and special symbols from the input text data. To vectorize the data we have used Tf-Idf Vectorizer from scikit-learn library. Tf-Idf vectorizer converts the input text data to a matrix of TF-IDF features (max features). This explains the second block of Figure 2. The generated matrix from the Tfidf vectorizer is finally fed to SVM or the Multilayer Perceptron to classify the clinical notes.

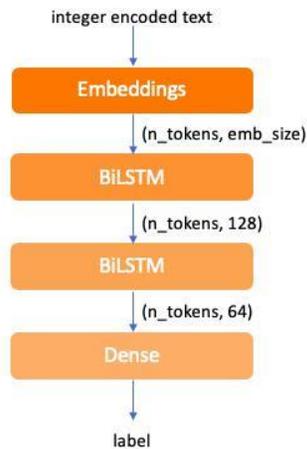

Fig. 1. The Deep Learning model using Word Embeddings and BiLstm.

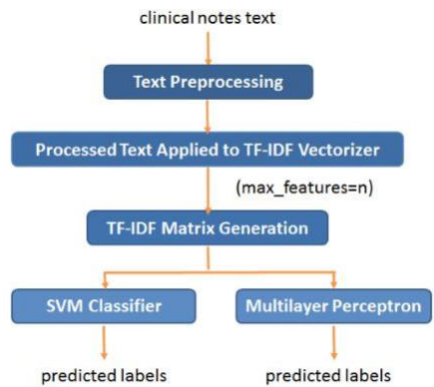

Fig. 2. Flow chart of the Support Vector Machine and the Multilayer Perceptron model used with the TF-IDF.

### 4.5 Experimental Setup

The Deep Learning model described in Section 4.4 was fed with all 3 data representation models introduced in Section 4.3. The model was trained by using the *rmsprop*[9] optimizer. The number of epochs we used is 20. This number is due to the fact that the resulting datasets do not have a great number of records and the convergence of the model is very fast. Finally, in order to prevent biased performances of our model, given to a non-balanced distribution of the records, we re-sampled our data by using the 10 fold cross-validation procedure. The performances were measured by $F_1$ score provided by scikit-learn[10] library.

## 5 Results And Discussion

In this section we report the results of our experiments and discuss about var-ious representation methods we used to capture domain peculiarities. Table 2 reports the results that were obtained by the Deep Learning model fed by the Word Embedding representations and SVM and Multilayer Perceptron TF-IDF models. TF-IDF approach was a state-of-the-art method for many years before the advent of Word Embeddings for textual knowledge representation for Ma-chine Learning algorithms. Thus, to compare our approach with two baselines that do not employ Deep Learning technologies, we decided to use TF-IDF and SVM model and TF-IDF and Multilayer Perceptron model.

As far as the comparison of Pre-trained Word2Vec and GloVe is concerned, it comes up that the latter performs better for all morbidity classes among the deep learning approaches. This indicates that GloVe Word Embeddings enable the Deep Learning model to better recognize diseases that are described within the input clinical notes with respect to the other deep learning models. This is in contrast with previous works where Word2Vec algorithm performed better for tasks of classification.

Moreover, although in literature domain-specific Word Embeddings proved to outperform pre-trained embeddings (e.g., in e-learning domain [6]), in this paper we have noticed the other way around. This happens because the dataset that we used to train the domain-specific embeddings is not big enough to allow to the algorithm to learn domain peculiarities and, hence, state-of-the-art embeddings, trained on a lot of texts, are still the best option among the deep learning methods.

From the results presented in Table 2, it turns out that both the baselines beat all the deep learning approaches. Reason of that lies behind specific features that appear alone for categories. When this happens the classical machine learn-ing algorithms are able to perform the classification with very high precision thanks to way the feature vector has been built using the TF-IDF technique.

---

[9] https://keras.io/optimizers/
[10] https://scikit-learn.org/

Table 2. Performance evaluation of the employed Deep Learning and Non-Deep Learn-ing models

| Morbidity class | Pre-trained GloVe F₁ | Pre-trained Word2Vec F₁ | Domain Word2Vec F₁ | Tf-Idf SVM F₁ | Tf-Idf Multilayer Perceptron F₁ |
|---|---|---|---|---|---|
| *Asthma* | 93.45 | 84.57 | 57.51 | **99.16** | 97.37 |
| *CAD* | 90.50 | 73.05 | 54.40 | **98.94** | 96.60 |
| *CHF* | 93.85 | 89.25 | 56.74 | **99.12** | 98.12 |
| *Depression* | 94.83 | 87.60 | 60.54 | **96.90** | 96.80 |
| *Diabetes* | 94.80 | 89.72 | 56.90 | 97.35 | **98.57** |
| *Gallstones* | 85.96 | 75.65 | 54.00 | **99.00** | 98.69 |
| *GERD* | 82.17 | 63.59 | 54.49 | **97.44** | 96.74 |
| *Gout* | 91.43 | 69.63 | 53.96 | 99.40 | **100** |
| *Hypercholesterolemia* | 90.56 | 88.02 | 66.05 | **97.96** | 96.56 |
| *Hypertension* | 95.92 | 82.80 | 57.29 | **97.66** | 97.03 |
| *Hypertriglyceridemia* | 93.92 | 72.57 | 61.77 | **99.40** | 98.98 |
| *OA* | 89.33 | 70.02 | 57.75 | **98.29** | 96.68 |
| *Obesity* | 85.49 | 64.19 | 48.72 | **97.42** | 95.91 |
| *OSA* | 94.77 | 83.92 | 55.38 | **98.59** | 99.50 |
| *PVD* | 98.25 | 81.55 | 55.86 | **98.93** | 98.19 |
| *Venous Insufficiency* | 84.23 | 66.16 | 47.95 | **100** | 97.79 |
| *Average* | 91.21 | 77.64 | 56.20 | **98.47** | 97.72 |

## 6   Conclusions and Future Work

In this paper we have investigated a Deep Learning model and employed three Word Embedding representations in order to recognize morbidity within clinical notes. The preliminary results indicate that certain features make the classifica-tion strongly biased and classical machine learning approaches are thus able to perform efficiently the classification. They are also able to outperform the Deep Learning strategies that have been adopted in combination with several word embeddings. These results suggest us the next directions we need to explore: identify the set of features for each category with the highest importance and that make thus the classification of machine learning methods using the TF-IDF feature engineering strategy easy to be performed. Once we discover these features, we will include one more action in the preprocessing step in order to make the dataset less biased. We will therefore repeat the comparisons with deep learning approaches and analyse again the results of the morbidity classification. As a further future work, we would like to study other Deep Learning models that might be morbidity-oriented. One more final direction worth to explore is the collection and pre-processing of further data. We remind that the used dataset consists of a few samples, where each of them includes a long text made of sentences and tokens. A possible strategy to employ would be to consider the different sentences as further samples each with the same annotation or, maybe, structure the text as a graph-shape and employ graph embeddings. Finally, we would also like to make experiments with other Word Embedding representations such as Bidirectional Encoder Representations from Transformers (BERT) [9].

## Acknowledgement

The research leading to these results has received funding from the EU's Marie Curie training network PhilHumans - Personal Health Interfaces Leveraging Human-Machine Natural interactions under grant agreement 812882. Moreover, we gratefully acknowledge the support of NVIDIA Corporation with the dona-tion of the Titan X GPU used for this research.